\documentclass[11pt]{article}
\usepackage{coling2016}
\usepackage{times}
\usepackage{url}
\usepackage{latexsym}
\usepackage{bm}
\usepackage{amssymb}
\usepackage{epsfig}
\usepackage{amsmath}
\usepackage{multirow}
\usepackage{graphicx}
\usepackage{caption}
\usepackage{subcaption}
\usepackage[titletoc,toc,title]{appendix}

\newcommand{\dotproduct}{\raisebox{.3ex}{\tiny$\bullet$}} 

\title{Toward Multilingual Neural Machine Translation \\with Universal Encoder and Decoder}

\author{Thanh-Le Ha, Jan Niehues, Alexander Waibel \\
 Institute for Anthropomatics and Robotics \\
KIT - Karlsruhe Institute of Technology, Germany \\
{\small \tt firstname.lastname@kit.edu}
}

\date{}

\begin{document}
\maketitle
\begin{abstract}
In this paper, we present our first attempts in building a multilingual Neural Machine Translation framework under a unified approach. We are then able to employ attention-based NMT for many-to-many multilingual translation tasks. Our approach does not require any special treatment on the network architecture and it allows us to learn minimal number of free parameters in a standard way of training. Our approach has shown its effectiveness in an under-resourced translation scenario with considerable improvements up to 2.6 BLEU points. In addition, the approach has achieved interesting and promising results when applied in the translation task that there is no direct parallel corpus between source and target languages.  
\end{abstract}

\section{Introduction}
\label{intro}
Neural Machine Translation (NMT) has shown its effectiveness in translation tasks when NMT systems perform best in recent machine translation campaigns~\cite{cettolo2015iwslt,bojar2016findings}. Compared to phrase-based Statistical Machine Translation (SMT) which is basically an ensemble of different features trained and tuned separately, NMT directly modeling the translation relationship between source and target sentences. Unlike SMT, NMT does not require much linguistic information and large monolingual data to achieve good performances.

An NMT consists of an encoder which recursively reads and represents the whole source sentence into a context vector and a recurrent decoder which takes the context vector and its previous state to predict the next target word. It is then trained in an end-to-end fashion to learn parameters which maximizes the likelihood between the outputs and the references. Recently, attention-based NMT has been featured in most state-of-the-art systems. First introduced by \cite{Bahdanau2014}, attention mechanism is integrated in decoder side as feedforward layers. It allows the NMT to decide which source words should take part in the predicting process of the next target words. It helps to improve NMTs significantly. Nevertheless, since the attention mechanism is specific to a particular source sentence and the considering target word, it is also specific to particular language pairs. 

Some recent work has focused on extending the NMT framework to multilingual scenarios. By training such network using parallel corpora in number of different languages, NMT could benefit from additional information embedded in a common semantic space across languages.  Basically, the proposed NMT are required to employ multiple encoders or multiple decoders to deal with multilinguality. Furthermore, in order to avoid the tight dependency of the attention mechanism to specific language pairs, they also need to modify their architecture to combine either the encoders or the attention layers. These modifications are specific to the purpose of the tasks as well. Thus, those multilingual NMTs are more complicated, much more free parameters to learn and more difficult to perform standard trainings compared to the original NMT.

In this paper, we introduce a unified approach to seamlessly extend the original NMT to multilingual settings.  Our approach allows us to integrate any language in any side of the encoder-decoder architecture with only one encoder and one decoder for all the languages involved. Moreover, it is not necessary to do any network modification to enable attention mechanism in our NMT systems. We then apply our proprosed framework in two demanding scenarios: under-resourced translation and zero-resourced translation. The results show that bringing multilinguality to NMT helps to improve individual translations. With some insightful analyses of the results, we set our goal toward a fully multilingual NMT framework. 

The paper starts with a detailed introduction to attention-based NMT. In Section~\ref{related}, related work about multi-task NMT is reviewed. Section~\ref{proposed} describes our proposed approach and thorough comparisons to the related work. It is followed by a section of evaluating our systems in two aforementioned scenarios, in which different strategies have been employed under a unified approach (Section \ref{evaluation}). Finally, the paper ends with conclusion and future work.

%
\blfootnote{
    %
    %
    %
    %
    %
    
     \hspace{-0.65cm}  
     This work is licenced under a Creative Commons 
     Attribution 4.0 International License.
     License details:
     \url{http://creativecommons.org/licenses/by/4.0/}
}

\section{Neural Machine Translation: Background}
\label{NMT}


An NMT system consists of an encoder which automatically learns the characteristics of a source sentence into fix-length context vectors
and a decoder that recursively combines the produced context vectors with the previous target word to 
generate the most probable word from a target vocabulary.

More specifically, a bidirectional recurrent encoder reads every words $x_{i}$ of a source sentence $\bm{x}=\{x_1,...,x_n\}$ 
and encodes a representation $\bm{s}$ of the sentence into a fixed-length vector $\bm{h}_i$ 
concatinated from those of the forward and backward directions:

\[
\begin{aligned}
& \bm{h}_i=[\overrightarrow{\bm{h}}_i,\overleftarrow{\bm{h}}_i] \\
& \overrightarrow{\bm{h}}_i=f(\overrightarrow{\bm{h}}_{i-1},\bm{s}) \\
& \overleftarrow{\bm{h}}_i=f(\overleftarrow{\bm{h}}_{i+1},\bm{s}) \\
& \bm{s}=\bm{E}_s~\dotproduct~\bm{x}_i \\
\end{aligned}
\]

Here $\bm{x}_i$ is the one-hot vector of the word $x_i$ and $\bm{E}_s$ is the word embedding matrix which is shared across the source words. 
$f$ is the recurrent unit computing the current hidden state of the encoder based on the previous hidden state. $\bm{h}_i$ is then called an \textit{annotation vector}, 
which encodes the source sentence up to the time $i$ from both forward and backward directions. Recurrent units in NMT can be a simple recurrent neural network unit (RNN), a Long Short-Term Memory unit (LSTM)~\cite{Hochreiter1997} or a Gated Recurrent Unit (GRU)~\cite{Cho2014}

Similar to the encoder, the recurrent decoder generates one target word $y_{j}$ to form a translated target sentence $\bm{y}=\{y_1,...,y_m\}$ in the end. 
At the time $j$, it takes the previous hidden state of the decoder $\bm{z}_{j-1}$, the previous embedded word representation $\bm{t}_{j-1}$ 
and a time-specific context vector $\bm{c}_j$ as inputs to calculate the current hidden state $\bm{z}_{j}$:

\[
\begin{aligned}
& \bm{z}_{j}=g(\bm{z}_{j-1}, \bm{t}_{j-1}, \bm{c}_j) \\
& \bm{t}_{j-1} = \bm{E}_t~\dotproduct~\bm{y}_{j-1}
\end{aligned}
\]

Again, $g$ is the recurrent activation function of the decoder and $\bm{E}_t$ is the shared word embedding matrix of the target sentences. 
The context vector $\bm{c}_j$ is calculated based on the annotation vectors from the encoder. 
Before feeding the annotation vectors into the decoder, an \textit{attention mechanism} is set up in between, 
in order to choose which annotation vectors should contribute to the predicting decision of the next target word. 
Intuitively, a relevance between the previous target word 
and the annotation vectors can be used to form some attention scenario.
There exists several ways to calculate the relevance as shown in ~\cite{Luong2015b}, but what we describe here follows the proposed method of ~\cite{Bahdanau2014}
\[
\begin{aligned} \label{eq:1}
& rel\_sc(\bm{z}_{j-1},\bm{h}_i)) = \bm{v}_a~\dotproduct~\tanh(\bm{W}_a~\dotproduct~\bm{z}_{j-1} + \bm{U}_a~\dotproduct~\bm{h}_i) \\
& \alpha_{ij} = \displaystyle  \frac{\exp(rel\_sc(\bm{z}_{j-1},\bm{h}_i))}{\sum_{i'} \exp(rel\_sc(\bm{z}_{j-1},\bm{h}_{i'}))},~\bm{c}_j = \displaystyle \sum_{i}{\alpha_{ij}\bm{h}_i}\\
\end{aligned}
\]

In ~\cite{Bahdanau2014}, this attention mechanism, originally called \textit{alignment model}, 
has been employed as a simple feedforward network with the first layer is a learnable layer via $\bm{v}_a$,$\bm{W}_a$ and $\bm{U}_a$.
The relevance scores $rel\_sc$ are then normalized into attention weights $\alpha_{ij}$ 
and the context vector $\bm{c}_j$ is calculated as the weighted sum of all annotation vectors $\bm{h}_i$. 
Depending on how much attention the target word at time $j$ put on the source states $\bm{h}_i$, a soft alignment is learned.   
By being employed this way, word alignment is not a latent variable but a parametrized function, making the alignment model differentiable. Thus, 
it could be trained together with the whole architecture using backpropagation. 
%


One of the most severe problems of NMT is handling of the rare words, 
which are not in the short lists of the vocabularies, i.e. out-of-vocabulary (OOV) words, or do not appear in the training set at all.
In~\cite{Luong2015a}, the rare target words are copied from their aligned source words after the translation.
This heuristic works well with OOV words and named entities but unable to translate unseen words.
In~\cite{Sennrich2016a}, their proposed NMT models have been shown to not only 
be effective on reducing vocabulary sizes but also have the ability to generate unseen words. 
This is achieved by segmenting the rare words into subword units and translating them. 
The state-of-the-art translation systems essentially employ subword NMT~\cite{Sennrich2016a}.


\section{Universal Encoder and Decoder for Multilingual Neural Machine Translation}
\label{multiling}
While the majority of previous research has focused on 
improving the performance of NMT on individual language pairs with individual NMT systems, recent work has
started investigating potential ways to conduct the translation involved in multiple languages using a single NMT system.
The possible reason explaining these efforts lies on the unique architecture of NMT. 
Unlike SMT, NMT consists of separated neural networks for the source and target sides, or the encoder and decoder, respectively.
This allows these components to map a sentence in any language to a representation in an embedding space 
which is believed to share common semantics 
among the source languages involved\footnote{But not necessarily syntactics since the embeddings 
are learned from parallel sentences which essentially share the same meaning although they might be very different in word order}.
From that shared space, the decoder, with some implicit or explicit relevant constraints, could transform the representation into a concrete sentence 
in any desired language. In this section, we review some related work on this matter. 
We then describe a unified approach toward an universal attention-based NMT scheme. Our approach does not require any architecture modification and it can be trained to learn a minimal number of parameters compared to the other work.
\subsection{Related Work}
\label{related}
By extending the solution of sequence-to-sequence modeling using encoder-decoder architectures to multi-task learning, \newcite{Luong2016} managed to achieve better performance on some $many-to-many$ tasks such as translation, parsing and image captioning compared to individual tasks. Specifically in translation, the work utilizes multiple encoders to translate from multiple languages, and multiple decoders to translate to multiple languages. In this view of multilingual translation, each language in source or target side is modeled by one encoder or decoder, depending on the side of the translation. Due to the natural diversity between two tasks in that multi-task learning scenario, e.g. translation and parsing, it could not feature the attention mechanism although it has proven its effectiveness in NMT. 

There exists two directions which proposed for multilingual translation scenarios where they leverage the attention mechanism. The first one is indicated in the work from \cite{Dong2015}, where it introduce an \textit{one-to-many} multilingual NMT system to translates from one source language into multiple target languages. Having one source language, the attention mechanism is then handed over to the corresponding decoder. The objective function is changed to adapt to multilingual settings. In testing time, the parameters specific to a desired language pair are used to perform the translation.

\newcite{Firat2016} proposed another approach which genuinely delivers attention-based NMT to multilingual translation.  As in \cite{Luong2016}, their approach utilizes one encoder per source language and one decoder per target language for  \textit{many-to-many} translation tasks. Instead of a quadratic number of independent attention layers, however, one single attention mechanism is integrated into their NMT, performing an affine transformation between the hidden layer of $m$ source languages and that one of $n$ target languages. It is required to change their architecture to accomodate such a complicated shared attention mechanism.

In a separate effort to achieve multilingual NMT, the work of \newcite{Zoph2016} leverages available parallel data from other language pairs to help reducing possible ambiguities in the translation process into a single target language\footnote{An example taken from the paper is when we want to translate the English word \textit{bank} into French, it might be easier if we have an additional German sentence containing the word \textit{Flussufer} (\textit{river bank}).  }. They employed the multi-source attention-based NMT in a way that only one attention mechanism is required despite having multiple encoders. To achieve this, the outputs of the encoders were combined before feeding to the attention layer. They implemented two types of encoder combination; One is adding a non-linear layer on the concatenation of the encoders' hidden states. The other is using a variant of LSTM taking the respective gate values from the individual LSTM units of the encoders. As a result, the combined hidden states contain information from both encoders , thus encode the common semantic of the two source languages.                        


\subsection{Universal Encoder and Decoder}
\label{proposed}
Inspired by the multi-source NMT as additional parallel data in several languages are expected to benefit single translations, we aim to develop a NMT-based approach toward an universal framework to perform multilingual translation. Our solution features two treatments: 1) Coding the words in different languages as different words in the language-mixed vocabularies and 2) Forcing the NMT to translating a representation of source sentences into the sentences in a desired target language. \\

\textbf{Language-specific Coding.}~~When the encoder of a NMT system considers words across languages as different words, with a well-chosen architecture, it is expected to be able to learn a good representation of the source words in an embedding space in which words carrying similar meaning would have a closer distance to each others than those are semantically different. This should hold true when the words have the same or similar surface form, such as (\textit{\textbf{@de@}Obama}; \textit{\textbf{@en@}Obama}) 
or (\textit{\textbf{@de@}Projektion}; \textit{\textbf{@en@}projection})\footnote{\textit{\textbf{@lang\_code@}a\_word} is a simple way 
that transforms the word \textit{a\_word} into a different surface form associated with its language \textit{\textbf{lang\_code}}. 
For example, \textit{\textbf{@de@}Projektion} is referred to the word \textit{Projektion} appearing in a German (\textit{\textbf{de}}) sentence.}. 
This should also hold true when the words have the same or similar meaning across languages, such as (\textbf{@en@}\textit{car}; \textit{\textbf{@en@}automobile}) 
or (\textit{\textbf{@de@}Flussufer}; \textit{\textbf{@en@}bank}). Our encoder then acts similarly to the one of multi-source approach\cite{Zoph2016}, collecting additional information from other sources for better translations, but with a much simpler embedding function. Unlike them, we need only one encoder, so we could reduce the number of parameters to learn. Furthermore, we neither need to change the network architecture nor depend on which recurrent unit (GRU, LSTM or simple RNN) is currently using in the encoder.

We could apply the same trick to the target sentences and thus enable \textit{many-to-many} translation capability of our NMT system. Similar to the multi-target translation\cite{Dong2015}, we exploit further the correlation in semantics of those target sentences across different languages. The main difference between our approach and the work of \cite{Dong2015} is that we need only one decoder for all target languages. Given one encoder for multiple source languages and one decoder for multiple target languages, it is trivial to incorporate the attention mechanism as in the case of a regular NMT for single language translation. In training, the attention layers were directed to learn relevant alignments between words in specific language pair and forward the produced context vector to the decoder.         
Now we rely totally on the network to learn good alignments between source and target sides. In fact, giving more information, our system are able to form nice alignments. 

In comparison to other research that could perform complete multi-task learning, e.g. the work from \cite{Luong2016} or the approach proposed by \cite{Firat2016}, our method is able to accommodate the attention layers seemlessly and easily. It also draws a clear distinction from those works in term of the complexity of the whole network: considerably less parameters to learn, thus reduces overfitting, with a conventional attention mechanism and a standard training procedure.   \\
 
\textbf{Target Forcing.}~~While language-specific coding allows us to implement a multilingual attention-based NMT, 
there are two issues we have to consider before training the network. 
The first is that the number of rare words would increase in proportion with the number of languages involved. 
This might be solved by applying a rare word treatment method with appropriate awareness of the vocabularies' size. 
The second one is more problematic: 
Ambiguity level in the translation process definitely increases due to the additional introduction of words having the same or similar meaning across languages
at both source and target sides. 
We deal with the problem by explicitly forcing the attention and translation to the direction that we prefer, 
expecting the information would limit the ambiguity to the scope of one language instead of all target languages. 
We realize this idea by adding at the beginning and at the end of every source sentences a special symbol indicating the language they would be translated into. 
For example, in a multilingual NMT, when a source sentence is German and the target language is English, the original sentence (already language-specific coded) is: \\
{\tt @de@darum @de@geht @de@es @de@in @de@meinem @de@Vortrag} \\
Now when we force it to be translated into English, the target-forced sentence becomes:\\
{\tt <E> @de@darum @de@geht @de@es @de@in @de@meinem @de@Vortrag <E>}

Due to the nature of recurrent units used in the encoder and decoder, in training, those starting symbols\footnote{For a bidirectional encoder, they are actually the starting symbols of a source sentence from two directions.} encourage the network learning the translation of following target words in a particular language pair. In testing time, information of the target language we provided help to limit the translated candidates, hence forming the translation in the desired language.
   
\begin{figure}[!htp]
\centering
  \includegraphics[width=0.9\columnwidth]{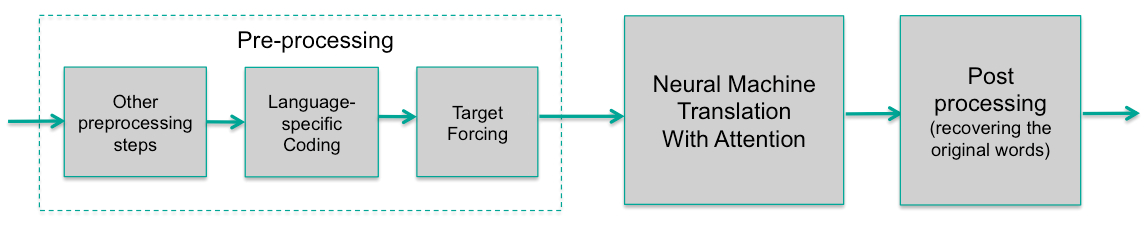}
\caption{\label{fig:steps} {\it Preprocessing steps to employ a multilingual attention-based NMT system}}
\end{figure}

Figure~\ref{fig:steps} illustrates the essence of our approach. With two steps in the preprocessing phase, namely language-specific coding and target forcing, we are able to employ multilingual attention-based NMT without any special treatment in training such a standard architecture. Our encoder and attention-enable decoder can be seen as a shared encoder and decoder across languages, or an \textit{universal} encoder and {decoder}. The flexibitily of our approach allow us to integrate any language into source or target side. As we will see in Section~\ref{evaluation}, it has proven to be extremely helpful not only in low-resourced scenarios but also in translation of well-resourced language pairs as it provides a novel way to make use of large monolingual corpora in NMT.  

\section{Evaluation}
\label{evaluation}
In this section, we describe the evaluation of our proposed approach in comparisons with the strong baselines using NMT in two scenarios:
the translation of an under-resource language pair and the translation of a language pair that does not exist any paralled data at all.

\subsection{Experimental Settings}
~~~~\textbf{Training Data.}~~We choose WIT3's TED corpus~\cite{cettolo2012} as the basis of our experiments since it might be the only high-quality parallel data of many low-resourced language pairs.
TED is also multilingual in a sense that it includes numbers of talks which are commonly translated into many languages.  
%
In addition, we use a much larger corpus provided freely by WMT organizers\footnote{\url{http://www.statmt.org/wmt15/}}
when we evaluate the impact of our approach in a real machine translation campaign. It includes the paralled corpus extracted from the digital corpus of European Parliament (EPPS),
the News Commentary (NC) and the web-crawled parallel data (CommonCrawl). While the number of sentences in popular TED corpora varies from 13 thousands to 17 thousands, the total number of sentences in those larger corpus is approximately 3 million sentences.  \\


\textbf{Neural Machine Translation Setup.}~~All experiments have been conducted using NMT framework {\tt Nematus}\footnote{\url{https://github.com/rsennrich/nematus}}, Following the work of \newcite{Sennrich2016a},
subword segmentation is handled in the prepocessing phase using Byte-Pair Encoding (BPE).
Excepts stated clearly in some experiments, we set the number of BPE merging operations at 39500 on the joint of source and target data.
When training all NMT systems, we take out the sentence pairs exceeding 50-word length and shuffle them inside every minibatch.
Our short-list vocabularies contain 40,000 most frequent words while the others are considered as rare words and applied the subword translation.
We use an 1024-cell GRU layer and 1000-dimensional embeddings with dropout at every layer
with the probability of 0.2 in the embedding and hidden layers and 0.1 in the input and ourput layers.
We trained our systems using gradient descent optimization with Adadelta~\cite{zeiler2012adadelta} on minibatches of size 80 and the gradient is rescaled whenever its norm exceed 1.0.
All the trainings last approximately seven days if the early-stopping condition could not be reached.
At a certain time, an external evaluation script on BLEU~\cite{papineni2002bleu} is conducted  on a development set to decide the early-stopping condition.
This evaluation script has also being used to choose the model archiving the best BLEU on the development set
instead of the maximal loglikelihood between the translations and target sentences while training.
In translation, the framework produces $n$-best candidates and we then use a beam search with the beam size of 12 to get the best translation.
\subsection{Under-resourced Translation}
First, we consider the translation for an under-resourced pair of languages.
Here a small portion of the large parallel corpus for English-German is used as a simulation for the scenario where we do not have much parallel data: Translating texts in English to German.
We perform language-specific coding in both source and target sides. 
By accommodating the German monolingual data as an additional input (German$\rightarrow$German), which we called the \textit{mix-source} approach,
we could enrich the training data in a simple, natural way.
Given this under-resourced situation, it could help our NMT obtain a better representation of the source side,
hence, able to learn the translation relationship better.
Including monolingual data in this way might also improve the translation of some rare word types such as named entities.
Furthermore, as the ultimate goal of our work, we would like to investigate the advantages of multilinguality in NMT.
We incorporate a similar portion of French-German parallel corpus into the English-German one. As discussed in Section~\ref{proposed}, it is expected to help reducing the ambiguity in translation between one language pair since it utilizes the semantic context provided by the other source language. We name this \textit{mix-multi-source}.

\begin{figure}
    \centering
    \begin{subfigure}[b]{0.35\textwidth}
        \includegraphics[width=\textwidth]{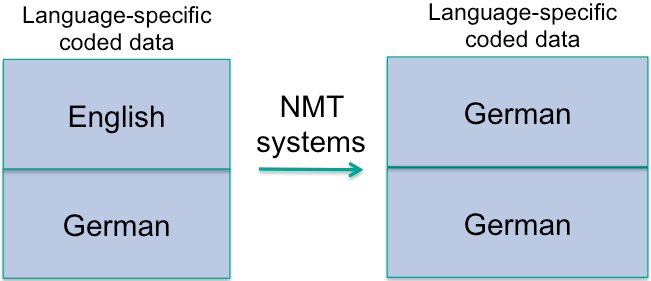}
        \caption{\textit{mix-source} system}
        \label{fig:gull}
    \end{subfigure}
    \qquad \qquad ~~~ 
    \begin{subfigure}[b]{0.35\textwidth}
        \includegraphics[width=\textwidth]{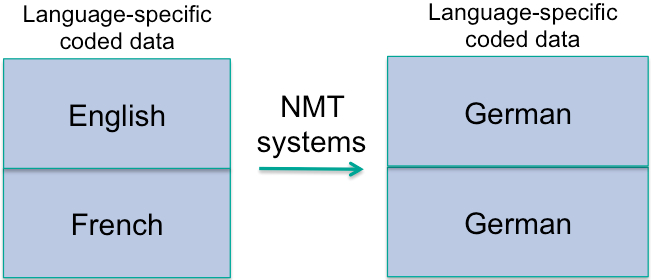}
        \caption{\textit{mix-multi-source} system}
        \label{fig:tiger}
    \end{subfigure}
    \caption{Different strategies of multi-source NMT in under-resourced translation}\label{fig:animals}
\end{figure}

\begin{table} [h!] 
\label{table:underresourced}
\centerline{ 
\begin{tabular}{|l|c|c|c|c|}
\cline{1-5}
\multirow{2}{*}{\textbf{System}} & \multicolumn{2}{c|}{ \textbf{tst2013}} & \multicolumn{2}{c|}{ \textbf{tst2014}}  \\ 
\cline{2-5}
 &
BLEU & $\Delta$BLEU & BLEU & $\Delta$BLEU    \\ \hline \hline
Baseline (En$\rightarrow$De)  & 24.35 & -- & 20.62 & --  \\ \hline
\cline{1-5}
Mix-source (En,De$\rightarrow$De,De)  & 26.99 & +2.64 & 22.71 & +2.09  \\ 
Mix-multi-source (En,Fr$\rightarrow$De,De) & 26.64 & +2.21 & 22.21 & +1.59  \\ \hline
\cline{1-5}
\end{tabular}}
\caption{\label{table:underresourced} {\textit{Results of the English$\rightarrow$German systems in an under-resourced scenario.}}}
\end{table}

Table~\ref{table:underresourced} summarizes the performance of our systems measured in BLEU\footnote{We used the script {\tt mteval-v13a.pl} of the Moses framework (\url{http://statmt.org/moses/}) as the official way to calculate BLEU scores in main machine translation campaigns.} on two test sets, \textit{tst2013} and \textit{tst2014}. Compared to the baseline NMT system which is solely trained on TED English-German data, our \textit{mix-source} system achieves a considerable improvement of 2.6 BLEU points on \textit{tst2013} and 2.1 BLEU points on and \textit{tst2014} . Adding French data to the source side and their corresponding German data to the target side in our \textit{mix-multi-source} system also help to gain 2.2 and 1.6 BLEU points more on \textit{tst2013} \textit{tst2014}, respectively. We observe a better improvement from our \textit{mix-source} system compared to our \textit{mix-multi-source} system. We speculate the reason that the \textit{mix-source} encoder utilize the same information shared in two languages while the \textit{mix-multi-source} receives and processes similar information in the other language but not necessarily the same.  We might validate this hypothesis by comparing two systems trained on a common English-German-French corpus of TED. We put it in our future work's plan.

\begin{figure}[!htp]
\centering
\includegraphics[width=1\columnwidth]{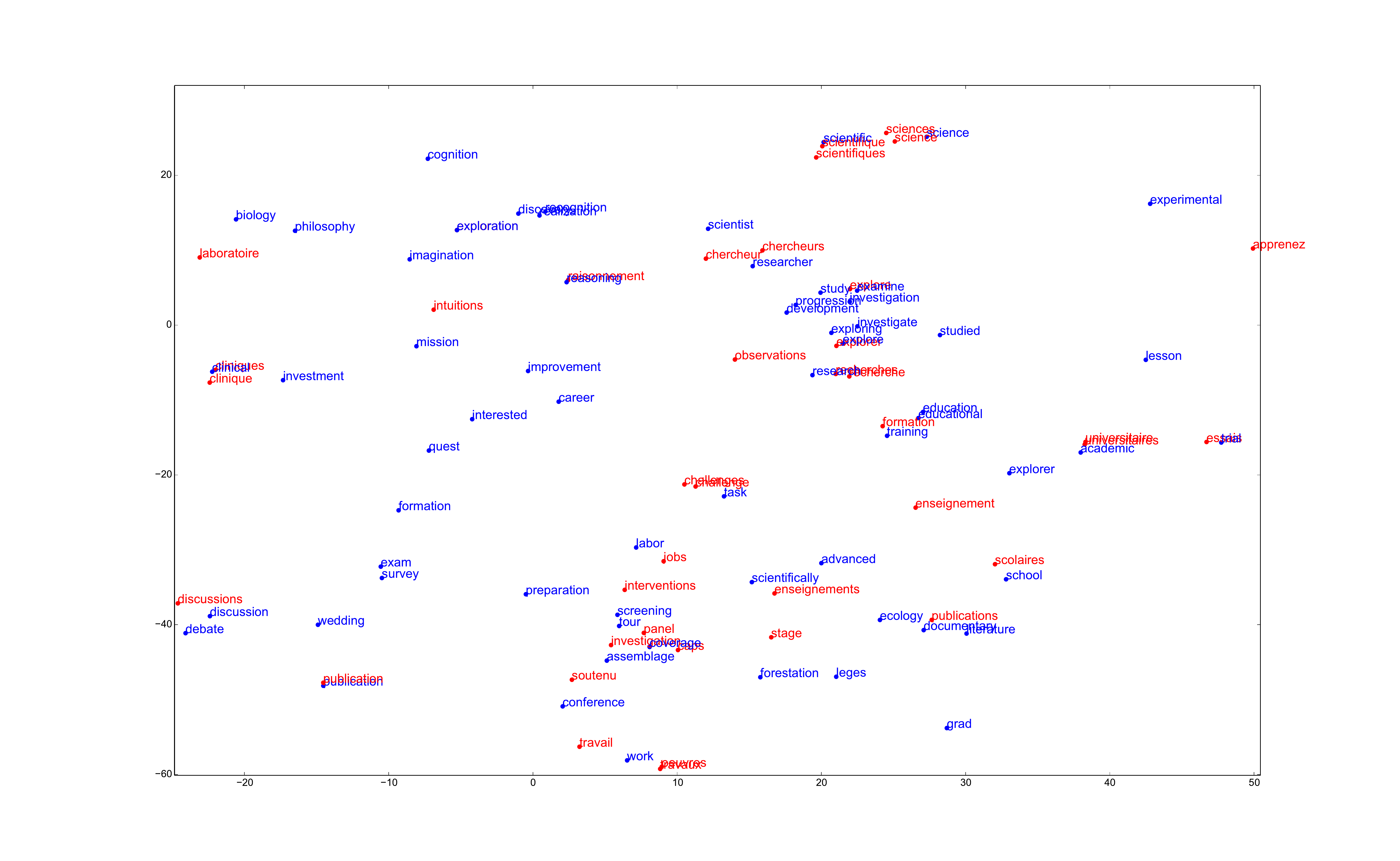}
\caption{\label{fig:MWE} {\it The multilingual word embeddings from the shared representation space of the source.}}
\vspace{-0.5cm}
\end{figure}

As we expected 
Figure~\ref{fig:MWE} shows how different words in different languages can be close in the shared space after being learned to translate into a common language. 
We extract the word embeddings from the encoder of the \textit{mix-multi-source} (En,Fr$\rightarrow$De,De) after training, 
remove the language-specific codes (\textit{\textbf{@en@}} and \textit{\textbf{@fr@}})and project the word vectors to the 2D space using t-SNE\footnote{To illustrate more clearly, only the word vectors of the words related to ``research'' are projected and visualized. 
The blue words are the English words and the red ones are the French words.}\cite{maaten2008visualizing}. 

\subsection{Using large monolingual data in NMT.}
A standard NMT system employs parallel data only. While good parallel corpora are limited in number, getting monolingual data of an arbitrary language is trivial. To make use of German monolingual corpus in an English$\rightarrow$German NMT system, ~\newcite{sennrich2016b} built a separate German$\rightarrow$English NMT using the same parallel corpus, then they used that system to translate the German monolingual corpus back to English, forming a synthesis parallel data.~\newcite{gulcehre2015} trained another RNN-based language model to score the monolingual corpus and integrate it to the NMT system through shallow or deep fusion. Both methods requires to train separate systems with possibly different hyperparameters for each. Conversely, by applying \textit{mix-source} method to the big monolingual data, we need to train only one network. We mix the TED parallel corpus and the substantial monolingual corpus (EPPS+NC+CommonCrawl) and train a \textit{mix-source} NMT system from those data. 

The first result is not encouraging when its performance is even worse than the baseline NMT which is trained on the small parallel data only. Not using the same information in the source side, as we discussed in case of \textit{mix-multi-source} strategy, could explain the degrading in performance of such a system. But we believe that the magnitude and unbalancing of the corpus are the main reasons. The data contains nearly four millions sentences but only around twenty thousands of them (0.5\%) are the genuine parallel data. As a quick attempt, after we get the model with that big data, we continue training on the real parallel corpus for some more epochs. When this adaptation is applied, our system brings an improvement of +1.52 BLEU on \textit{tst2013} and +1.06 BLEU on \textit{tst2014} (Table~\ref{table:monoNMT}).   

\begin{table} [h!] 
\label{table:monoNMT}
\centerline{ 
\begin{tabular}{|l|c|c|c|c|}
\cline{1-5}
\multirow{2}{*}{\textbf{System}} & \multicolumn{2}{c|}{ \textbf{tst2013}} & \multicolumn{2}{c|}{ \textbf{tst2014}}  \\ 
\cline{2-5}
 &
BLEU & $\Delta$BLEU & BLEU & $\Delta$BLEU    \\ \hline \hline
Baseline (En$\rightarrow$De)  & 24.35 & -- & 20.62 & --  \\ \hline
\cline{1-5}
Mix-source big (En,De$\rightarrow$De,De)  & 25.87 & +1.52 & 21.68 & +1.06   \\ 
\cline{1-5}
\end{tabular}}
\caption{\label{table:monoNMT} {\textit{Results of the English$\rightarrow$German system using large monolingual data.}}}
\end{table}

\subsection{Zero-resourced Translation}
Among low-resourced scenarios, zero-resourced translation task stands in an extreme level. A zero-resourced translation task is one of the most difficult situation when there is no parallel data between the translating language pair. To the best of our knowledge, there have been yet existed a published work about using NMT for zero-resourced translation tasks up to now. In this section, we extend our strategies using the proposed multilingual NMT approach as first attempts to this extreme situation.

We employ language-specific coding and target forcing in a strategy called \textit{bridge}. Unlike the strategies used in under-resourced translation task, \textit{bridge} is an entire \textit{many-to-many} multilingual NMT. Simulating a zero-resourced German$\rightarrow$French translation task given the available German-English and English-French parallel corpora, after applying language-specific coding and target forcing for each corpus, we mix those data with an English-English data as a ``bridge'' creating some connection between German and French. We also propose a variant of this strategy that we incorporate French-French data. And we call it \textit{universal}. 

We evaluate \textit{bridge} and \textit{universal} systems on two German$\rightarrow$French test sets. They are compared to a \textit{direct} system, which is an NMT trained on German$\rightarrow$French data, and to a \textit{pivot} system, which essentially consists of two separate NMTs trained to translate from German to English and English to French. The \textit{direct} system should not exist in a real zero-resourced situation. We refer it as the perfect system for comparison purpose only. In case of the \textit{pivot} system, to generate a translated text in French from a German sentence, we first translate it to English, then the output sentence is fed to the English$\rightarrow$German NMT system to obtain the French translation. Since there are more than two languages involved in those systems, we increase the number of BPE merging operations proportionally in order to reduce the number of rare words in such systems. We do not expect our proposed systems to perform well with this primitive way of building direct translating connections since this is essentially a difficult task. 
We report the performance of those systems in Table~\ref{table:zeroresourced}.

\begin{table} [h!] 
\label{table:zeroresourced}
\centerline{ 
\begin{tabular}{|l|c|c|} \hline
System & BLEU & $\Delta$BLEU    \\ \hline
Direct (De$\rightarrow$Fr)  & 16.65 & +3.24  \\ 
Pivot (De$\rightarrow$En$\rightarrow$Fr)  & 13.41 & --   \\ 
Bridge (De,En,En$\rightarrow$En,Fr,En) & 9.70 & -3.71  \\ 
Universal (De,En,En,Fr$\rightarrow$En,Fr,En,Fr) & 10.77 & -2.64 \\ \hline
\end{tabular}}
\caption{\label{table:zeroresourced} {\textit{Results of the German$\rightarrow$French systems in a zero-resourced scenario.}}}
\end{table}

Unsupprisingly, both \textit{bridge} and \textit{universal} systems perform worse than the \textit{pivot} one. We consider two possible reasons:

\textbf{Our target forcing mechanism is moderately primitive.}~~Since the process is applied after language-specific coding, the target forcing symbol is the same for all source sentences in every languages. Thus, the forcing strength might not be enough to guide the decision of the next words. Once the very first word is translated into a word in wrong language, the following words tend to be translated into that wrong language again. Table~\ref{table:zrstats} shows some statistics of the translated words and sentences in wrong language.   

 \begin{table} [h!] 
\label{table:zrstats}
\centerline{ 
\begin{tabular}{|l|c|c|} \hline
 System & \% Translated words in wrong language & \% Sentences in wrong language    \\ \hline
Bridge & 21.27\% & 9.70\%  \\ 
Universal & 17.57 & 9.47\% \\ \hline
\end{tabular}}
\caption{\label{table:zrstats} {\textit{Percentages of language identificcation mistakes when applying our translation strategies.}}}
\end{table}

\textbf{Balancing of the training corpus.}~~Although it is not severe as in the case of \textit{mix-source} system for large monolingual data, the limited number of sentences in target language can affect the training. The difference of 1.07 BLEU points between \textit{bridge} and \textit{universal} might explain this assumption as we added more target data (French) in \textit{universal} strategy, thus reducing the unbalance in training. 
        
Those issues would be addressed in our following future work toward the multilingual attention-based NMT.
\section{Conclusion and Future Work}
In this paper, we present our first attempts in building a multilingual Neural Machine Translation framework. By treating words in different languages as different words and force the attention and translation to the direction of desired target language, we are able to employ attention-enable NMT toward a multilingual translation system. Our proposed approach alleviates the need of complicated architecture re-designing when accommodating attention mechanism. In addition, the number of free parameters to learn in our network does not go beyond that magnitute of a single NMT system. With its universality, our approach has shown its effectiveness in an under-resourced translation task with considerable improvements. In addition, the approach has achieved interesting and promising results when applied in the translation task that there is no direct parallel corpus between source and target languages.

Nevertheless, there are issues that we can continue working on to address in future work. A more balancing data would be helpful for this framework. The mechanism of forcing the NMT system to the right target language could be improved. We could conduct more detailed analyses of the various strategies under the framework to show its universarity.

%

\bibliographystyle{acl}
\bibliography{coling2016}

\end{document}